\documentclass[preprint,12pt]{elsarticle}

\usepackage{amssymb}
\usepackage{float}
\usepackage{amsthm}

\journal{Journal}

\begin{document}

\begin{frontmatter}



\title{Mask Usage Recognition using Vision Transformer with Transfer Learning and Data Augmentation}


\author{Hensel D. Jahja}
\author{Novanto Yudistira\corref{cor1}}
\ead{yudistira@ub.ac.id}
\author{Sutrisno}

\affiliation{organization={Informatics Department, Faculty of Computer Science, Brawijaya University},
            addressline={Jalan Veteran 8, Malang, 65145}, 
            city={Malang, },
            country={Indonesia}}

\begin{abstract}
The COVID-19 pandemic has disrupted various levels of society. The use of masks is essential in preventing the spread of COVID-19 by identifying an image of a person using a mask. Although only 23.1\% of people use masks correctly, Artificial Neural Networks (ANN) can help classify the use of good masks to help slow the spread of the Covid-19 virus. However, it requires a large dataset to train an ANN that can classify the use of masks correctly. MaskedFace-Net is a suitable dataset consisting of 137016 digital images with 4 class labels, namely Mask, Mask Chin, Mask Mouth Chin, and Mask Nose Mouth. Mask classification training utilizes Vision Transformers (ViT) architecture with transfer learning method using pre-trained weights on ImageNet-21k, with random augmentation. In addition, the hyper-parameters of training of 20 epochs, an Stochastic Gradient Descent (SGD) optimizer with a learning rate of 0.03, a batch size of 64, a Gaussian Cumulative Distribution (GeLU) activation function, and a Cross-Entropy loss function are used to be applied on the training of three architectures of ViT, namely Base-16, Large-16, and Huge-14.
Furthermore, comparisons of with and without augmentation and transfer learning are conducted. This study found that the best classification is transfer learning and augmentation using ViT Huge-14. Using this method on MaskedFace-Net dataset, the research reaches an accuracy of 0.9601 on training data, 0.9412 on validation data, and 0.9534 on test data. This research shows that training the ViT model with data augmentation and transfer learning improves classification of the mask usage, even better than convolutional-based Residual Network (ResNet).
\end{abstract}




\end{frontmatter}


\section{Introduction}
\label{sec:introduction}
During the COVID-19 pandemic, the economy, medical resources, and various types of sectors have disrupted the health and development of the affected communities. At the time when COVID-19 first appeared, the two countries most affected (China and South Korea) recommended the use of masks to reduce the spread of coronavirus 2 (SARS-CoV-2) \citep{Feng2020}.
The use of masks was controversial in several countries at the beginning of coronavirus 2 first known. However, previous studies in respiratory diseases such as H1N1 influenza have shown significant results in reducing the spread of the virus by using face masks \citep{Cowling2010}. In addition, research on the risk assessment of the spread of coronavirus 2, has shown that wearing a full-face mask can delay the spread of influenza \citep{Brienen2010}.
People who use masks can reduce virus transmission, but if masks are not properly used, it can increase the risk of spreading coronavirus 2 (World Health Organization, 2020). However, in a study conducted in Japan within its society, only 23.1\% of people used masks properly \citep{Machida2020}. Therefore, supervision of the use of masks is necessary because regions that carry out mandatory mask regulations have seen a decrease in cases infected with coronavirus 2 \citep{VanDyke2020}.

Artificial Neural Networks (ANN) can help classify the usage of the mask through image recognition by learning the extracted feature of various images by processing the images through the layer repeatedly. ANN will deliver the best performance incrementally with the size and uniqueness of the data set \citep{shahinfar2020datasize}. Through some research, we found that Masked Face-Net data set \citep{Cabani2021} is the best choice for this research. It consists of 137016 digital images using masks with 4 class labels: Mask, Mask Chin, Mask Mouth Chin, and Mask Nose Mouth. Compared to the data set used by \cite{wang2020RMFD} Masked Face Detection Dataset (MFDD) which consists of only 24771 masked face images, and Real-world Masked Face Recognition (RMFRD) which consist of only 95000 pictures, Masked Face-net has more range of variation of age, race, and ethnicity because it is gathered from Flickr-Faces-HQ dataset \citep{karras2018gan}. 

Since the published paper on Gradient-based learning applied to document recognition \citep{lecun1998cnn}, the Convolutional Neural Network (CNN) has become the standard of image recognition. Since then, various papers have been published to beat the predecessor results on data set such as ImageNet \citep{deng2009imagenet}. Almost all the state-of-the-art paper is based on the backbone of convolutional neural networks such as Inception \citep{szegedey2014inception}, VGG \citep{simonyan15vgg}, ResNet \citep{he2015resnet} and the latest EfficientNet \citep{tan2019efficientnet} which beat all the previous CNN architecture. Throughout the years, the trend for image recognition has been going deeper on the layer used in training \citep{alzubaidi2021surveycnn}, which means it requires more computing resources, hence causing inefficient research in ANN study with limited resources. Another research conducted by \citep{novanto2020} shows that the more layers in an ANN, will affect the training time. This will adversely affect the efficiency of computational time, especially for large data sets processing. However, Vision Transformer (ViT) \citep{Dosovitskiy2020}, another model architecture that has been proven to be faster is comparable to EfficientNet in performance while being 5 times faster in computation. ViT works using the attention mechanism used in paper Attention is all you need \citep{Vaswani2017} with positioning embedded flatten patches of images to attention mechanism and then passing through the multi-layer perceptron layer to classify each class.

Training an ANN might be complicated if we train from scratch. One method that can be applied is to use transfer learning. Transfer learning uses ANN weights, which have previously been trained on the larger data set, and fine-tuned on the unseen data set. Using this method improves accuracy compared to models created from scratch \citep{Barman2019}. The benchmark of pre-trained weights used for transfer learning is usually based on state-of-the-art dataset such as ImageNet21k \citep{deng2009imagenet}. ImageNet21k consists of more than 14 million images and 21 thousand categories. The enormous numbers of images and categories will help jump-start the learnable ANN parameter in the training process with immense accuracy to begin.
ANN performance improves concurrently with the size of the data set used in training \citep{alom2019datasetsize}. Moreover, data augmentation helps increase the accuracy and quality of the dataset by modifying the form of the data before being inserted into the model architecture. Furthermore, it improves the performance of deep neural networks (DNN) and thus increases the generalization of model \citep{Wang2016}.

\section{Related Works}
\label{sec:related_works}
Research has been conducted to classify masks to stay safe during the COVID-19 pandemic. Solutions to optimize mask usage recognition to prevent the spread of the virus are a hot topic. Research by \cite{Tomas2021} entitled "Incorrect Facemask-Wearing Detection Using CNN with Transfer Learning" uses a crowd data set with 13 categories. A manual label was carried out by a nursing group from the hospital of the Ontinyent, resulting in 3200 images from 500 users. To help boost the performance of this small data set, the authors compared multiple methods such as data augmentation and transfer learning. They evaluated the results of multiple architectures such as MobileNet \citep{howard2017mobilenets} that have the smallest size footprint of 3.5 million parameters compared to VGG16 \citep{simonyan15vgg} of 134.4 million parameters. VGG16 produces the best accuracy of 0.834 using transfer learning and data augmentation. Aside from the size of the model that can yield better performance, the usage of transfer learning and data augmentation always show better performance than those without.

Another research done by \cite{Loey2021maskdetection} used real-world masked face data set (RMFD) \citep{wang2020RMFD} which consists of 5000 masked and 90000 unmasked faces of real-world mask usage with similar faces. Labeled Faces in the Wild (LFW) \citep{kawulok2016advances} which consist of 13000 simulations of masked faces were introduced. Furthermore, they proposed the usage of a hybrid deep transfer learning model, with the ResNet50 (\cite{he2015resnet}) as a features extractor. Finally, the use of Support Vector Machines \citep{Cristianini2008svm} and decision tree \citep{breiman1984decisiiontree} to form an ensemble learning \citep{zhang2012ensemble} perform well with testing accuracy of 99.46\% on RMFD dataset and testing accuracy of 100\% on LFW data set.

Research by \cite{Bosheng2020} with the title "Identifying Facemask-Wearing Condition Using Image Super-Resolution with Classification Network to Prevent COVID-19" uses the Super-Resolution and Classification Networks (SRCNet) architecture, with transfer learning method capable of achieving up to 98.70\% accuracy. This research inspired our work to utilize transfer learning. Another paper has made annotated data set (\cite{Loey2021NovelDeep}) of Medical Masks Dataset (MMD), which consists of 853 images belonging to 3 categories of mask, without mask, and mask worn incorrectly. This yields much more precise result of 81\% by using YOLOv2 \citep{redmon2015yolo} and ResNet \citep{he2015resnet} feature extraction. The research done by \cite{Li2012annotated} concluded that probabilistic model training with annotated images should perform better than non-annotated images. However, the required number of annotated images in the deep learning research is hard to find and validate. The related works that we have mentioned inspire us to apply transfer learning and data augmentation to this research while using a much larger data set which is MaskedFace-Net, which should yield better performance, and be the first one to do so.

\section{Methods and solutions}
\label{sec:methods}
\subsection{Data set}\label{data_sets}
\begin{figure}[!h]
	\centering
	\includegraphics[scale=.4]{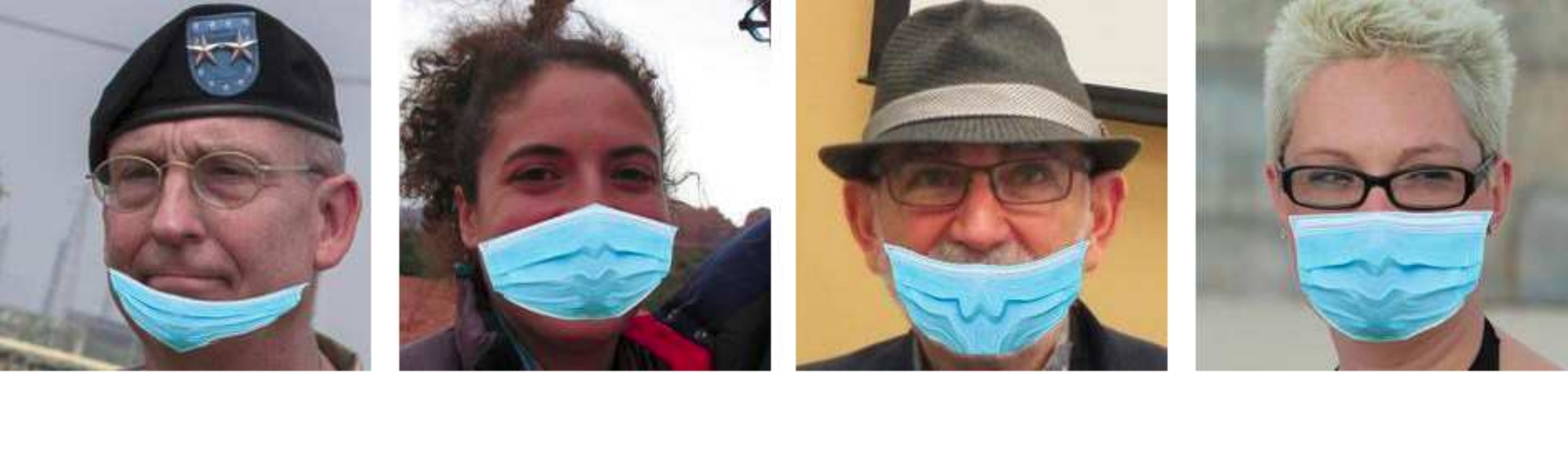}
	\caption{MaskedFace-Net sample.}
	\label{mfn_fig}
\end{figure}

The related works that we have cited in the previous chapter concluded that we need a large resolution and data set to achieve better accuracy, uniqueness of the images, and perform better in a real-world application.
The data that we used in this study was MaskedFace-Net, \citep{Cabani2021} which consists of 4 categories of Mask, Mask Chin, Mask Nose Mouth, and Mask Mouth Chin. The total of this data set is 137016 images with dimension of 1024 by 1024 pixels. The data set is widely sparse and non-biased to one category only, thus we believe this is the most suitable data set to be used in this research. The data set is based on Flicker \citep{karras2018gan} which is usually used to be training data for generative adversarial network (GAN) in image generation. Moreover, the author of MaskedFace-Net uses facial landmark detections to detect facial features and mask-to-face mapping to add a mask on the face. Fig. \ref{mfn_fig} shows samples from MaskedFace-Net. From left to right, we have Mask Chin category in the first column covered only the chin part of the face; Mask Nose Mouth, with mask usage that covered only the nose and the mouth; Mask Mouth Chin, which covered the mouth and the chin of the face, and; Mask category with mask usage that covers the nose, mouth, and chin;. Although it consists of only four categories, the data set is detailed, large, sparse, and thus capable of being used in various GAN \citep{karras2018gan}.

\subsection{Vision transformers}\label{vision_transformers}
Transformer was introduced in 2017 with a journal entitled Attention Is All You by \cite{Vaswani2017}, to overcome problems in the Recurrent Neural Network (RNN) \citep{Hopfield1982rnn} and Long Short Term Memory (LSTM) \citep{hocreiter1997lstm} models. However, both models have problems with loss gradients and long training times in the case of Natural Language Processing (NLP). For years, image classification tasks have always used CNN \citep{lecun1998cnn} as the backbone of the architecture. However, in paper by \cite{Dosovitskiy2020}, it was discovered that Transformers could be used for image classification research, with five times faster in computation time than the latest state-of-the-art convolutional architecture while keeping the accuracy head to head.

\begin{figure}[!h]
			\centering
			\includegraphics[scale=.35]{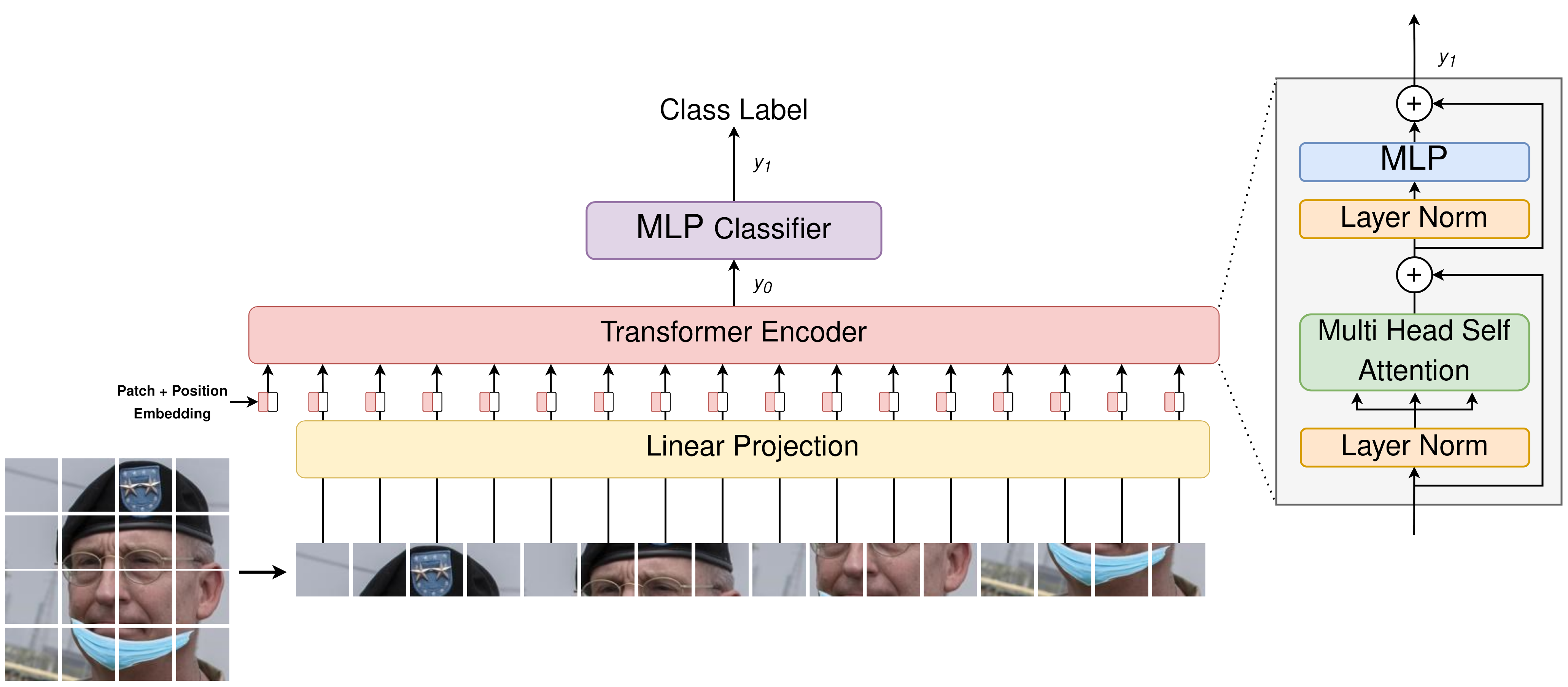}
			\caption{ViT Model Overview.}
			\label{ViT Fig.}
		\end{figure}

ViT is not the usual image classification architecture have been introduced before. Fig. \ref{ViT Fig.} shows an overview of the ViT model. Images are split into several parts based on the number of patches that have been declared after going through the process of splitting a 2-dimensional digital image. It is necessary to change the 2-dimensional digital image into a 1-dimensional vector. Eq. \ref{lp_eq} is a formula for changing a 2-dimensional image into a 1-dimensional vector, where \textit{H},\textit{W}, is the resolution of the image, and \textit{C} is the channels numbers. It then will be converted into a  
$R^{N\left(P^{2} \cdot C\right)}$, where $P$ is the number of patches and $N=HW/P^2$. After that, the embedding results will pass through the transformers encoder.
		ViT encoder behaves just like the encoder mechanism. We found Transformers for natural language processing task on the paper of Attention is All You Need \citep{Vaswani2017} in that it requires an embedded input. Then it is processed through layer normalization \cite{ba2016layer} which uses the distribution of the summed input to a neuron over a mini-batch of training cases. Mean and variance, which are then used to normalize the summed input, gives a huge time advantage compared to batch normalization. The multi-head self-attention is needed to capture the image's critical part. Eq. \ref{attn_eq} shows the equation for attention, where $Q$ is the query or the pure input value from the embedding, $K$ is the permutation of the input, and $V$ is the scaled dot product from $Q$ and $K$ with softmax activation. Multi-head attention contains the concatenation of multi self-attention. As seen in Eq. \ref{mha_eq} where the number of heads will be multiplied with the $W^{O}$ value. This gives the transformers encoder the best feature extraction to attend the important part. Unlike the other works that we have seen in Section \ref{sec:related_works}, most of the feature extraction uses ResNet \citep{he2015resnet} feature extractor to obtain the most critical part of the images. However, the self-attention on transformers encoder \citep{Dosovitskiy2020} will result well enough without another feature extraction. The last layer of transformers encoder is a simple multilayer perceptron, with each output is based on the category that we have defined in our data set, which in our case is four, with the activation of the multilayer perceptron utilizaing GeLU \citep{hendrycks2016gelu}.
		
		\begin{equation}
			x \in R^{HWC} \rightarrow x_{p} \in R^{N\left(P^{2}C\right)}
			\label{lp_eq}
		\end{equation}
		
		\begin{equation}
			AttentionHead(Q, K, V)=softmax(\frac{QK^T}{\sqrt{d_k}})V
			\label{attn_eq}
		\end{equation}
		
		\begin{equation}
			MultiHead(Q, K, V) = Concat(AttentionHead_1, ..., AttentionHead_n)W^O
			\label{mha_eq}
		\end{equation}
		
		\subsection{Data augmentation}
		The classification results of ANN becomes better when using a larger dataset \citep{shahinfar2020datasize} because the ANN learns from every pixel in a digital image. In addition, it can leverage the data augmentation process for data enrichment. data augmentation is a process using digital image processing, which changes images in such a way thus that transform those digital images as a new form of digital images \citep{Wang2016},  The benefit of data augmentation can also be seen in paper by \cite{fadhil2020} which shows that the use of augmentation has an effect on training outcomes of an ANN, by showing higher accuracy and lower loss values than those without data augmentation because data augmentation helps ANN recognize various patterns. There is much research in data augmentation methods to increase the performance of ANN. The latest state of the art is AutoAugment \citep{cubuk2018autoaugment} which applies augmentation of random choices on batch of images. AutoAugment works by using the searching algorithm controller Recurrent Neural Network, samples sum of data, and searching the probability of operation using the best result. The disadvantage of AutoAugment is that the process takes much time, especially in large data set. There is a RandAugment \citep{cubuk2019randaugment} solution that eliminates the search for the best augmentation in a phase so that the computation process is fast. By eliminating the search space for the base algorithm to classify the best results and using randomly distributed application augment on the whole data set, the RandAugment method reduce the search space from $10^{32}$ to $10^2$. Although it is much faster than AutoAugment, RandAugment yields the same accuracy results as the latest state-of-the-art, and it does not linearly increase the search space to the data set sizes. 
		
		
		\subsection{Transfer learning}
		The ViT architecture works best if the architecture has been trained using multiple data set before being fine-tuned on other data set \citep{Dosovitskiy2020}. The process of training the architecture using the pre-trained weights on each neuron, and tune it to another data set is called transfer learning. Transfer Learning will work well if the data set used in a transfer learning process has similarities, both from class, type of digital image to digital image resolution. However, the most influential factor on the success of transfer learning is the amount and diversity of data \citep{Weiss2016}. Therefore, this research will use a transformer vision architecture that has been trained on the ImageNet \citep{deng2009imagenet} data set, which is a data set containing 14197122 annotated images according to the WordNet hierarchy \citep{Russakovsky2015}.
		
		\subsection{Gradient-weighted class activation mapping (grad-cam)}
		Every application of ANN relied on human niche expertise to implement throughout the years. However, the explainability and comprehensibility of AI are necessary to be deployed in real-world domains because the user needs to understand the system works. Thus it can be adequately tested and referred \citep{yampolskiy2019unexplainable}.
		\cite{Selvaraju2020gradcam} proposed a technique for making ANN more explainable. Every training process using ANN needs a gradient to compute the update the weights. Gradient-weighted Class Activation Mapping uses this acquired gradient to produce a coarse localization map by highlighting the critical regions of an image on the last layer convolutional block.
		ViT work differently in that the architecture does not use any convolutional block. Instead, we will treat the last layer of the attention block that is not affected by token addition. Eq. \ref{pre-Gradcam Equation}  shows the computation of $w_k^{\left(c\right)}$ with the $H$ as the height of the image and $W$ represent the width of the image. This calculation is needed to sum up the matrix from the chosen layer gradient. Eq. \ref{Gradcam Equation} multiplies channel-wise Eq. \ref{pre-Gradcam Equation} with its activation before summing up. Finally, ReLU \citep{agarap2018relu} activation, which will return 0 on the value that less than 0  is utilized. This activation will eliminate unnecessary gradient to focus only on the most important part of the gradient mapping on the image.
		
		\begin{equation}
			w_k^{\left(c\right)}=\frac{1}{H\cdot W}\sum_{i=1}^{H}\sum_{j=1}^{W}\frac{\partial Y^{\left(c\right)}}{\partial A_k\left(i,j\right)}
			\label{pre-Gradcam Equation}
		\end{equation}
		
		\begin{equation}
			L_{Grad-CAM}^{\left(c\right)}\left(x,y\right)=ReLU\left(\sum_{k}{w_k^{\left(c\right)}A_k\left(x,y\right)}\right)
			\label{Gradcam Equation}
		\end{equation}
		\subsection{Model training}\label{model_training}
		There are various configurations for the ViT architecture. Constant hyperparameters will be set in training to prevent bias in the test results of each configuration. In this research, we will keep the same setup that  \cite{Dosovitskiy2020} used in the original ViT research. The hyperparameters used for fine-tuning are batch size of 64, a learning rate of 0.03, epochs of 20, loss function of Cross-Entropy \citep{zhang2018crossentropy}, optimizer Stochastic Gradient Descent, and  GeLU \citep{hendrycks2016gelu} activation. For this research, we split the training, validation, and test data set the size of 80\%, 10\%, and 10\%, respectively, from the whole data set of MaskedFaceNet \citep{Cabani2021}. For transfer learning, pre-trained weights that have been trained on ImageNet21K \citep{deng2009imagenet}, and augmentation method of RandAugment are utilized during training \citep{cubuk2019randaugment}. The hardware specification that is used throughout experiments are RTX 8000, Intel(R) Xeon(R) Gold 6230R, RAM 255 GiB. 

\section{Experimental Results}
		
		\subsection{Comparing Size of ViT}
		 Variants of ViT can be seen in Table \ref{vit_variants} with ViT Base 16 consist of 16x16 patches, 12 layers of transformers encoder, 768 hidden sizes, 3072 of multilayer perceptron in the transformers encoder, 12 attention heads, and number of parameters of 86 millions. ViT Large 16 consists of 16x16 patches, 24 layers of transformers encoder, 1024 hidden size, 4096 of multilayer perceptron in the transformers encoder, 16 attention heads, and the sum of parameters of 307 million. ViT Huge 14 consists of 14x14 patches, 32 layers of transformers encoder, 1280 hidden size, 5120 of multilayer perceptron in the transformers encoder, 16 attention heads, and the number of parameters of 632 million.
		We will discuss the effect of architectural size on the accuracy of training, validation, and test data. Table \ref{size_acc_table} shows the training results from 20 epochs, and the highest accuracy value will be taken for each architecture. The results show that the ViT Huge 14 yields the best results for all data set parts, with an impressive accuracy of 0.93 on the test set. Meanwhile, the transformers large 16 has the worse performance out of the three variants, despite having more than three times the size parameters of ViT Base 16.

		
				\begin{table*}[!h]
			\caption{ViT variants.}
			\label{vit_variants}
			\centering
			\begin{tabular}{p{2cm}|p{1cm}p{1cm}p{1cm}p{1cm}p{1cm}p{1.4cm}}
				Model & Patches & Layers & Hidden Size & MLP size & Heads & Num. of Params \\
				\hline
				ViT Base 16 &
				16 &
				12 &
				768 &
				3072 &
				12 &
				86 Mil.\\
				ViT Large 16 &
				16 &
				24 &
				1024 &
				4096 &
				16 &
				307 Mil.\\
				ViT Huge 14 &
				14 &
				32 &
				1280 &
				5120 &
				16 &
				632 Mil.
			\end{tabular}
		\end{table*}
		
\begin{figure}[H]
			\centering
			\includegraphics[scale=0.5]{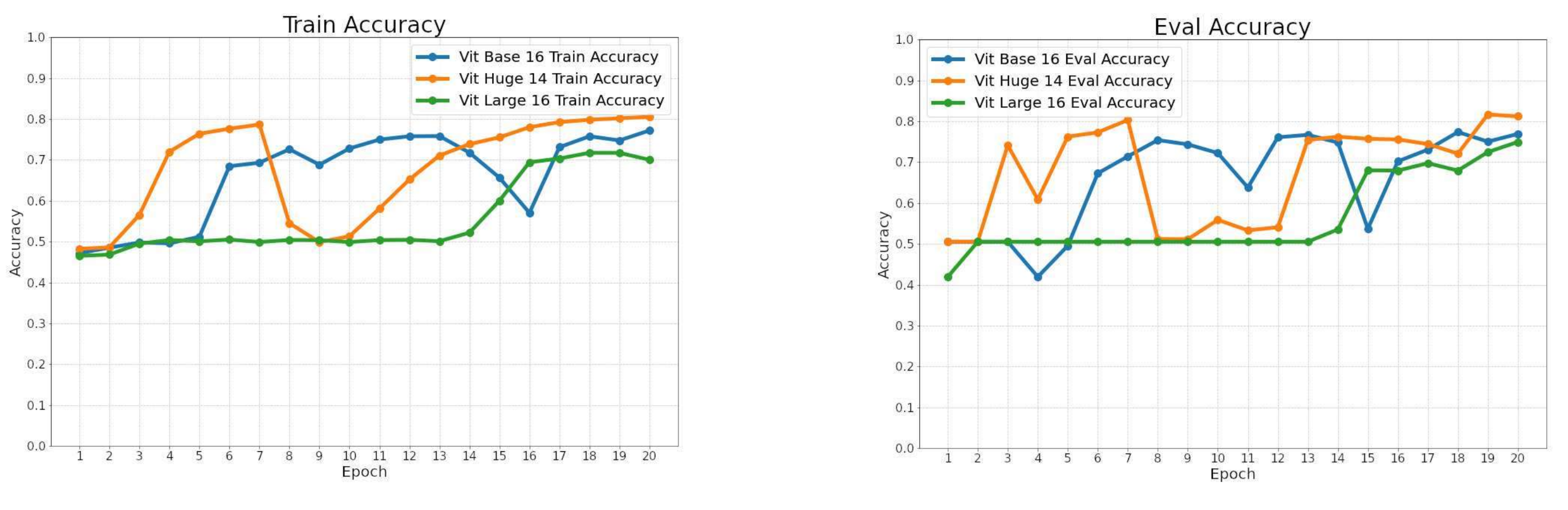}
			\caption{Accuracy of different ViT in 20 epochs}
			\label{size_acc_fig}
		\end{figure}

		The results of the ViT architecture training and validation epoch by epoch can be seen in Fig. \ref{size_acc_fig}. It shows that the highest accuracy was obtained by the ViT Huge 14 architecture at epoch of $20^{th}$ with an accuracy of 0.805519. A significant difference compared to other architectures, namely ViT Base 16 with accuracy of 0.772179, which was obtained at $14^{th}$ epoch. Finally, accuracy of ViT Large 16 of 0.717461 were obtained at $19^{th}$ epoch. The validation results for each ViT architecture did not differ much from the training results. In this case, the ViT Huge 14 got the highest accuracy at $19^{th}$ epoch with accuracy of 0.816742. This is higher than the ViT Base 16 architecture of 0.773756, which was obtained at $13^{th}$ epoch, and the ViT Large 16 architecture with accuracy of 0.749623 obtained at $20^{th}$ epoch. The evaluation results on the testing data show that the accuracy of ViT Huge 14 architecture is very high compared to the rest ViT architectures of 0.934586.
		
		\begin{table*}[!h]
			\caption{Accuracy of different ViT size}
			\label{size_acc_table}
			\centering
			\begin{tabular}{p{2cm}|p{2cm}p{2cm}p{2cm}}
				Model & Train & Validation & Test \\
				\hline
				ViT Base 16 &
				0.772179 &
				0.773756 &
				0.818045 \\
				ViT Large 16 &
				0.717461 &
				0.749623 &
				0.766917 \\
				ViT Huge 14 &
    			\textbf{0.805519} &
    			\textbf{0.816742} &
				\textbf{0.934586} \\
			\end{tabular}
		\end{table*}
		
		\subsection{Effects of augmentation on accuracy}
		Random Augment Policy will be used on training and validation data set considering the results of previous works. The ViT Huge 14 architecture will be used because this architecture yields the best accuracy on training, validation, and test set thus avoiding overfitting. In this test, we will compare the results of using and without augmentation, on training data with the same setup we have defined in Subsection \ref{model_training}
		
\begin{figure}[H]
			\centering
			\includegraphics[scale=.5]{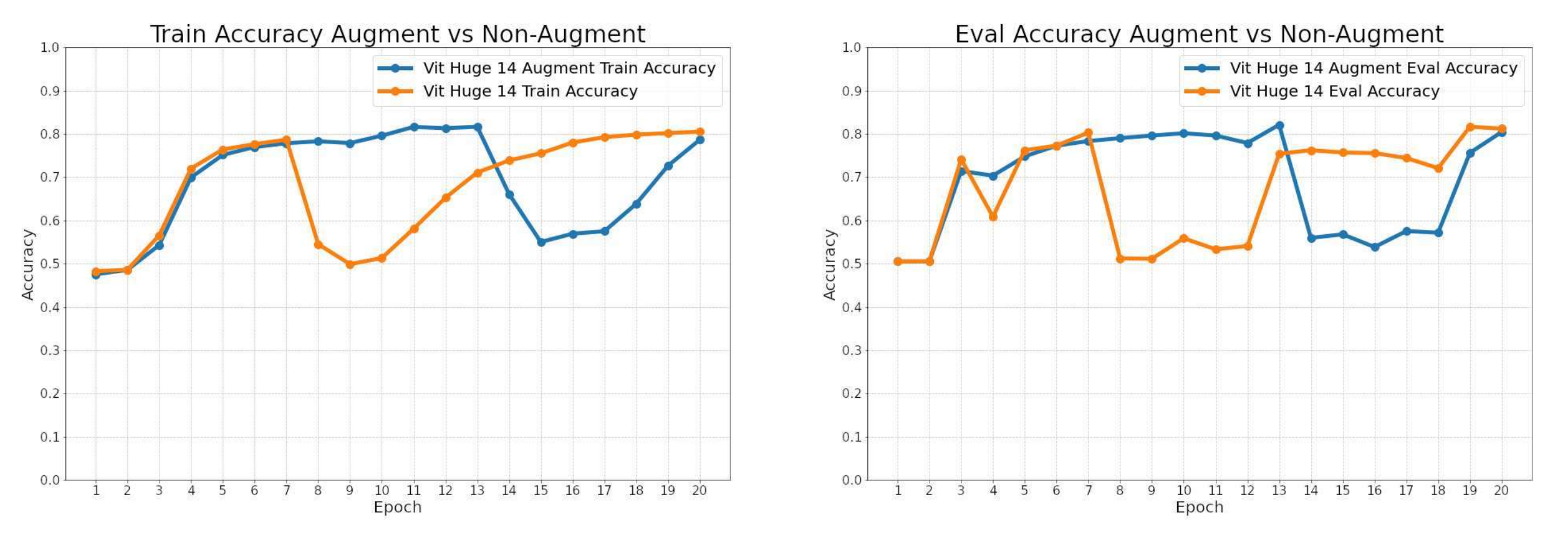}
			\caption{Impacts of augmentation on accuracy}
			\label{aug_acc_fig}
		\end{figure}

		From results on the accuracy of augmented and non-augmented data can be seen in Table \ref{aug_acc_table} where leveraging dataset with augmentation got higher accuracy, with a value of 0.816821 for training data, 0.82167 for validation data, compared to dataset without augmentation of 0.805519 for training data, and 0.816742 for validation data. A comparable accuracy is obtained in the test data, which is 0.934586. This indicates that the raw data which has been processed using augmentation, has comparable performance. The augmentation results on all set yield much better accuracy than without augmentation. Fig. \ref{aug_acc_fig} shows epoch by epoch accuracy on augmentation applied on ViT training, It shows that those ViT with augmentation yield the best results at the $13^{th}$ epoch, compared to training without the augmentation that yields the best results on the $20^{th}$ epoch. Validation on both augmented and non-augmented data set yields the same results as the training data set. We concluded that augmentation on the training converges faster than without augmentation.
		
		\begin{table*}[!h]
			\caption{Effects of augmentation on ViT Huge 14}
			\label{aug_acc_table}
			\centering
			\begin{tabular}{p{4cm}|p{2cm}p{2cm}p{2cm}}
				Model & Train & Validation & Test \\
				\hline
				Augmentation &
				\textbf{0.816821} &
				\textbf{0.821267} &
				\textbf{0.934586} \\
				Without augmentation &
				0.805519 &
				0.816742 &
				0.934586 \\
			\end{tabular}
		\end{table*}
		
		\subsection{Impacts of transfer learning method}
		
\begin{figure}[!h]
			\centering
			\includegraphics[scale=.5]{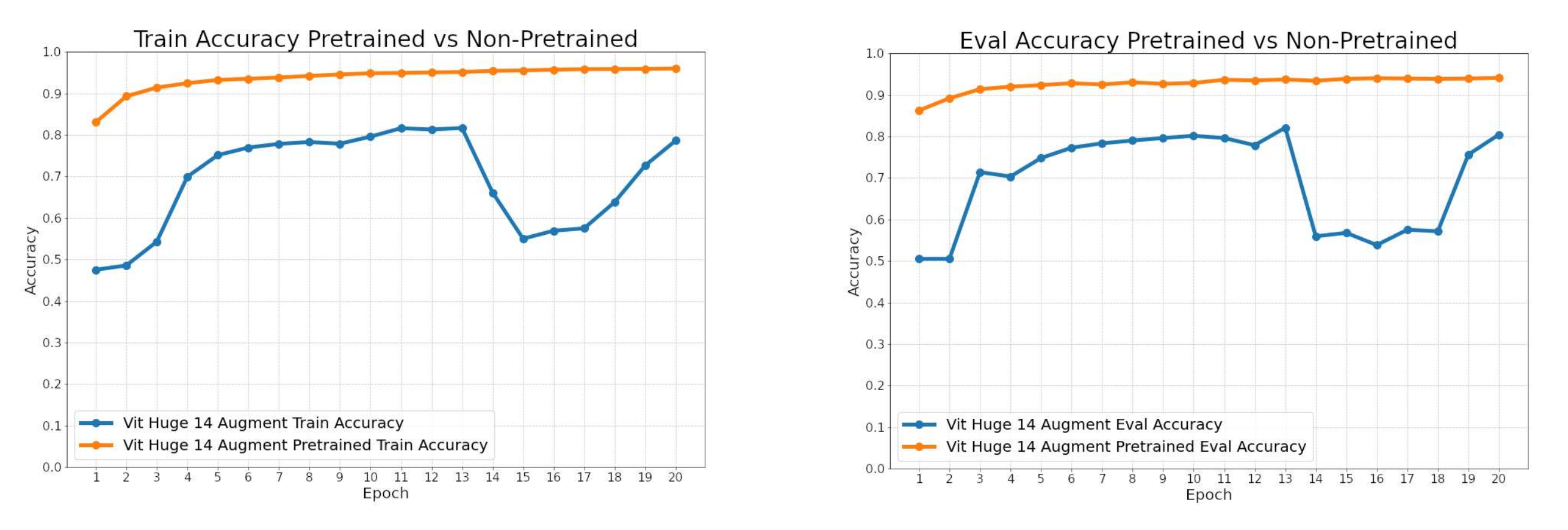}
			\caption{Impacts of transfer learning on accuracy}
			\label{tl_acc_fig}
		\end{figure}

		To evaluate the effect of pretrained weights, we compare the results of using pretrained weights and without using pretrained weights. In this subsection, we will use pretrained weights that have been trained using the ImageNet21K dataset on the ViT Huge 14 architecture, with training and validation data set that has been augmented. On testing dataset, the comparison of accuracy of the use of pretrained weights and without using pretrained weights are presented in Table \ref{tl_acc_table}. It shows that the best accuracy across all $20^{th}$ epochs that it has been trained. It was found that using pretrained weights got higher results with accuracy on training data of 0.960068, validation data of 0.941176, and on test data of 0.953383, compared to without using pretrained weights with accuracy on training data of 0,816821, validation data of 0.82167, and test data of 0.934586. Fig. \ref{tl_acc_fig} shows epoch by epoch of the training and validation phase. It shows that the model that uses pretrained weights starts the training with much better accuracy compared to without the usage of pre-trained weights. From epoch of $13^{th}$, there is a considerable drop in accuracy caused by gradient loss. It did not happen in the model with the pretrained weights. From these results, we can conclude that the usage of pretrained weights performs much better than without it.
		
		\begin{table*}[!h]
			\caption{Impacts of transfer learning on ViT Huge 14}
			\label{tl_acc_table}
			\centering
			\begin{tabular}{p{4cm}|p{2cm}p{2cm}p{2cm}}
				Model & Train & Validation & Test \\
				\hline
				With Transfer Learning &
				\textbf{0.960068} &
				\textbf{0.941176} &
				\textbf{0.953383}\\
				Without Transfer Learning &
				0.816821 &
				0.821267 &
				0.934586 \\
			\end{tabular}
		\end{table*}
		
		\subsection{Comparisons with baselines}
		After carrying out various experiments on augmentation and pretrained weights, it can be seen that the use of augmentation and pretrained weights simultaneously can increase accuracy and reduce losses in training, validation, and testing. Therefore, experiments using augmentation and pretrained weights will be carried out on all existing architectures and the residual network architecture as a benchmark in this study.

\begin{figure}[!h]
			\centering
			\includegraphics[scale=.5]{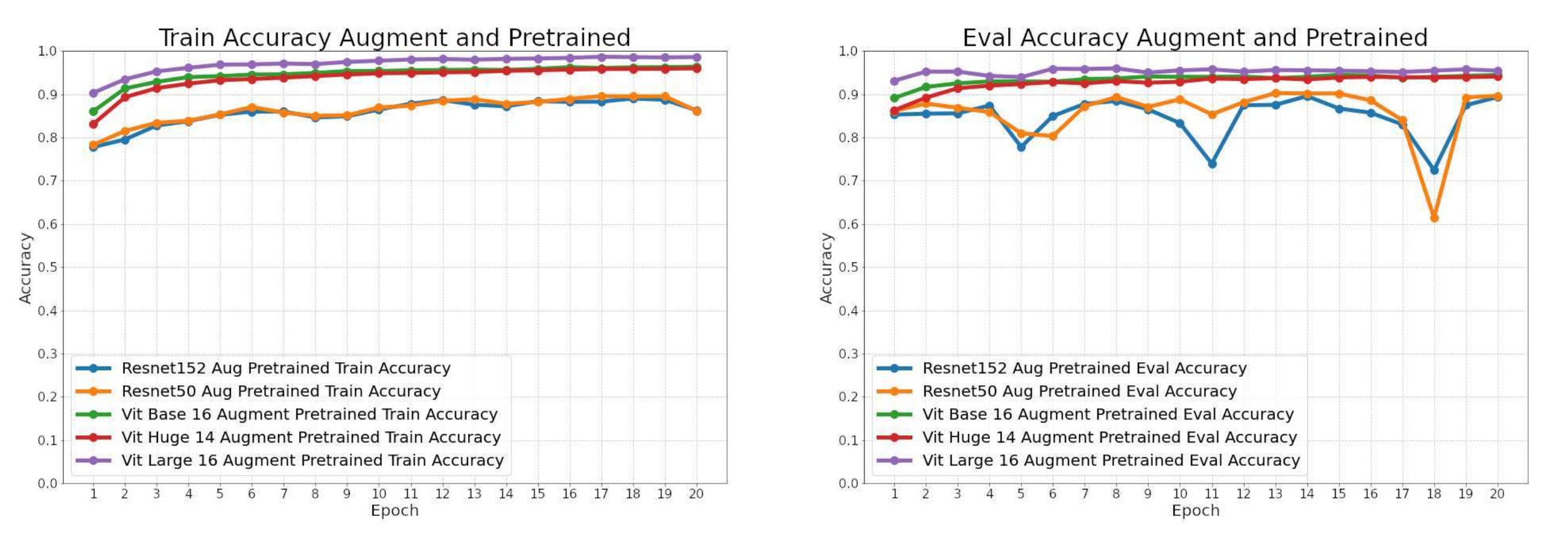}
			\caption{Impacts of augmentation and transfer learning on accuracy}
			\label{all_acc_fig}
		\end{figure}

		Testing results of each architecture can be seen on Table \ref{all_acc_table}. It was found that by using pretrained weights and data augmentation, the ViT Large 16 architecture produces the highest accuracy value on the training data of 0.986909, and the validation data of 0.960030. However, on the test data, it was found that the architecture ViT Huge 14 has an higher testing accuracy of 0.953383 compared to ViT Large 16 of 0.928571, which is evidence that there is overfitting problem for the architecture of ViT Large 16. Fig. \ref{all_acc_fig} shows the 20 epochs of training of all evaluated architectures. It can be revealed from the train and validation plots that ResNet \citep{he2015resnet} produces low performances on both validation and training. Furthermore, at $18^{th}$ epoch, there is a huge drop in accuracy caused by loss of gradient.
		
		\begin{table*}[h]
			\caption{Impacts of augmentation and transfer learning}
			\label{all_acc_table}
			\centering
			\begin{tabular}{p{3cm}|p{2cm}p{2cm}p{2cm}}
				Model & Train & Validation & Test \\
				\hline
				ViT Base 16 &
				0.963929 & 
				0.944947 &	
				0.820301\\
				ViT Large 16 &
				\textbf{0.986909} &
				\textbf{0.96003}	&
				0.928571\\
				ViT Huge 14 &
				0.960068 &
				0.941176 &
				\textbf{0.953383} \\
				ResNet 50 &
				0.672797 &
				0.521005 &
				0.553454 \\
				ResNet152 &
				0.792082 &
				0.697393&
				0.797244 \\
			\end{tabular}
		\end{table*}
		
		\subsection{Confusion matrix}
		The best test result was obtained by ViT Huge 14 using pretrained weights and augmentation. Table \ref{cm_table} shows the result of the confusion matrix on the test data. These tests show that the ViT Huge 14 with augmentation and pretrained weights can classify each class well. The Mask class prediction accuracy is 0.967543, Mask Chin is 0.925926, Mask Mouth Chin is 0.890701, and Mask Nose Mouth is 0.813953. The results of incorrectly presenting the usage seem to be low, with the highest rate of false postive is Mask Nose Mouth class classified as Mask of 0.162791.

		\begin{table*}[!h]
			\caption{Confusion matrix on every class}
			\label{cm_table}
			\centering
			\begin{tabular}{p{2.8cm}|p{1.5cm}p{1.5cm}p{1.5cm}p{1.5cm}}
				\hspace{2.5cm} & Mask & Mask Chin & Mask Mouth Chin & Mask Nose Mouth \\
				\hline
				Mask &
				0.967543 &
				0.000000 &
				0.012365 &
				0.020093 \\
				Mask Chin &
				0.000000 &
				0.925926 &
				0.074074 &
				0.000000 \\
				Mask Mouth Chin &
				0.061990 &
				0.032626 &
				0.890701 &
				0.014682 \\
				Mask Nose Mouth &
				0.162791 &
				0.000000 &
				0.023256 &
				0.813953 \\
			\end{tabular}
		\end{table*}

		\subsection{Grad-Cam Results}
\begin{figure}[H]
            \centering
			\includegraphics[scale=.6]{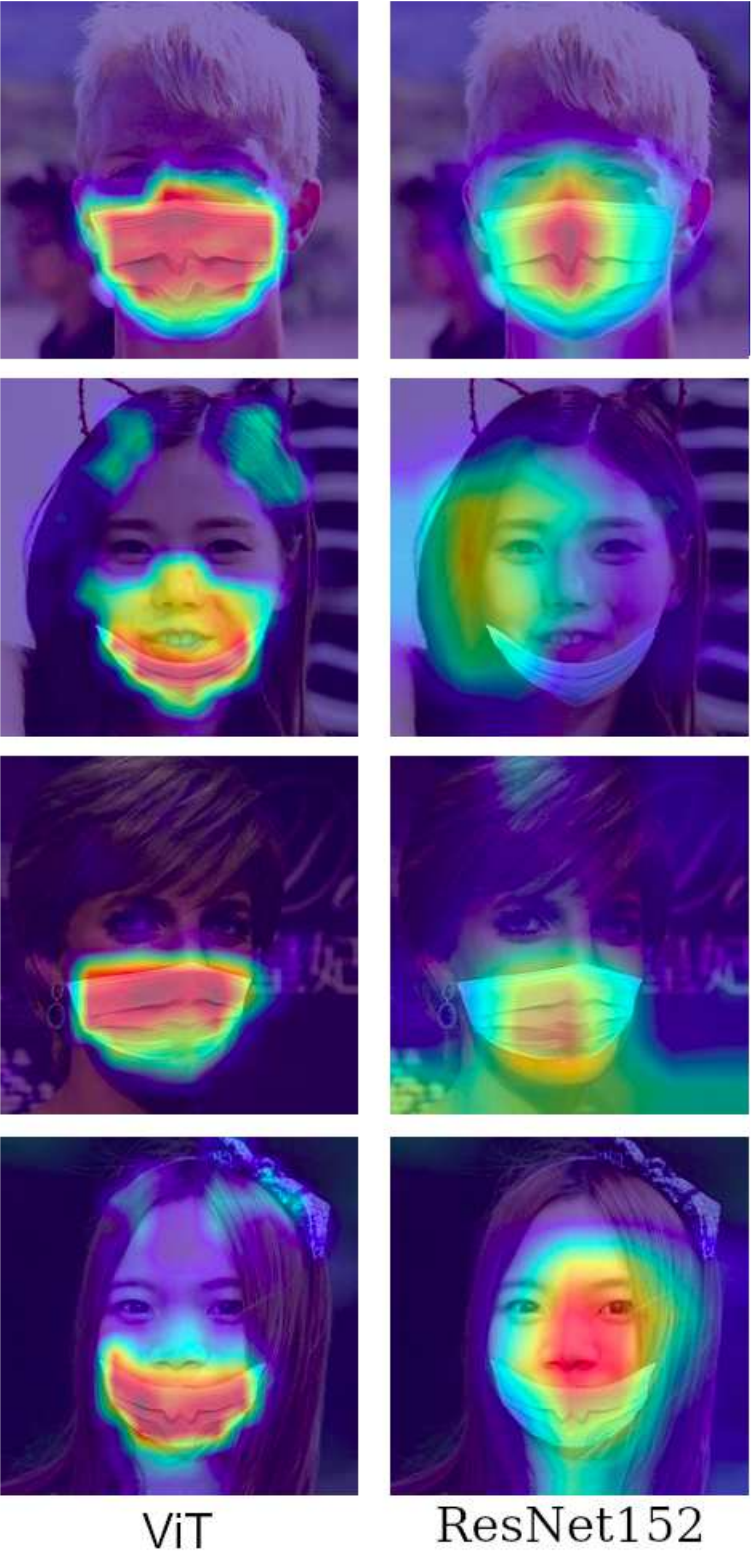}
			\caption{Grad-Cam results of ViT and ResNet152}
			\label{all_res}
		\end{figure}

		GradCam \citep{Selvaraju2020gradcam} requires a gradient on a layer so that the attention of a target layer is obtained. In ViT, we will use the gradient of the last attention layer in the architecture to find where the important parts are in an image. For the residual network architecture, we will use the last convolutional layer. Based on Table \ref{all_acc_table}, ResNet152  \citep{he2015resnet} performs better than ResNet50, which we will use as comparison with ViT Huge 14. Fig. \ref{all_res} shows the result of Grad-Cam for the Mask class. 
		
		Based on the visualization, it shows that ViT focuses on the entire mask area which is more appealing than ResNet152's attention where only focuses on the nose and mouth section behind the mask. The second row of figure of the Mask Chin class shows a significant difference between ViT and ResNet where ViT focuses on the part of the exposed face, while the gradient of ResNet152 does not focus on any part of the face. The third row shows the Grad-Cam result of the Mask Nose Mouth class. It can be revealed that the gradient of ViT much better focused on part of the mask, and looks more precise than ResNet152. The fourth row shows the Grad-Cam result of the Mask Mouth Chin class. It shows that the gradient of ViT focuses on the part of the mask, while the gradient of ResNet152 focus on the nose and eyes of the person.
		

		
		\section{Conclusion}
		
				Based on the research results on the classification of mask usage using the ViT architecture and data augmentation, it can be concluded that the best implementation is carried out by performing random augmentation, on each existing dataset. ViT architecture is then applied by processing the digital image according to the pre-determined number of patches. Then a linear projection reduces the digital image's shape from three to two dimensions. Later, embedding will be carried out before entering the encoder transformer, where attention will focus on the critical features in the digital image. The feature map eventually enters into the multi-layer perceptron for the last step before being classified according to the existing class. The accuracy of the classification mask usage using ViT and data augmentation produces 0.967543 for the Mask class, 0.925926 for the Mask Chin,  0.890701 for the Mask Mouth Chin, and 0.813953 for the Mask Nose Mouth class. Overall, the accuracy of 0.953383 shows that the proposed ViT on mask usage recognition outperforms the convolutional baseline method.


\bibliographystyle{elsarticle-num} 
\bibliography{cas-refs}




\end{document}